\documentclass[10pt,twocolumn,letterpaper]{article}

\usepackage{cvpr}              
\usepackage{graphicx}
\usepackage{amsmath}
\usepackage{amssymb}
\usepackage{booktabs}
\usepackage{hhline}
\usepackage{adjustbox}
\usepackage{multirow}

\usepackage{times}
\usepackage{epsfig}
\usepackage{graphicx,caption}
\usepackage{cuted}

\usepackage[pagebackref,breaklinks,colorlinks]{hyperref}

\usepackage[capitalize]{cleveref}
\crefname{section}{Sec.}{Secs.}
\Crefname{section}{Section}{Sections}
\Crefname{table}{Table}{Tables}
\crefname{table}{Tab.}{Tabs.}

\def\cvprPaperID{10601} 
\def\confName{CVPR}
\def\confYear{2022}

\begin{document}

\title{Controllable Animation of Fluid Elements in Still Images}

\author{Aniruddha Mahapatra\\
Adobe Research India\\
{\tt\small anmahapa@adobe.com}
\and
Kuldeep Kulkarni\\
Adobe Research India\\
{\tt\small kulkulka@adobe.com}
}
\maketitle

\begin{strip}
\centering
  \includegraphics[width=\textwidth]{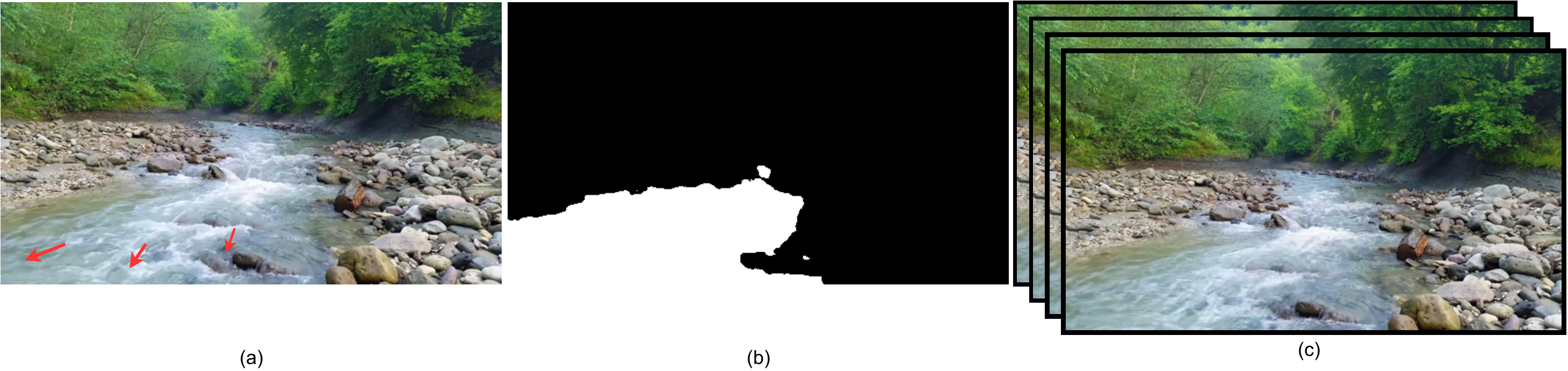}
  \captionof{figure}{Our approach takes in input image along with the user-provided motion hints (red arrows in (a) and the user-provided mask (white in (b)) indicating the regions of fluid elements to be animated and outputs the sequence of frames of the animated videos. }
\end{strip}

\begin{abstract}
   We propose a method to interactively control the animation of fluid elements in still images to generate cinemagraphs. Specifically, we focus on the animation of fluid elements like water, smoke, fire, which have the properties of repeating textures and continuous fluid motion. Taking inspiration from prior works, we represent the motion of such fluid elements in the image in the form of a constant 2D optical flow map. To this end, we allow the user to provide any number of arrow directions and their associated speeds along with a mask of the regions the user wants to animate. The user-provided input arrow directions, their corresponding speed values, and the mask are then converted into a dense flow map representing a constant optical flow map ($F_D$).  We observe that $F_D$, obtained using simple exponential operations can closely approximate the plausible motion of elements in the image. We further refine computed dense optical flow map $F_D$ using a generative-adversarial network (GAN) to obtain a more realistic flow map. We devise a novel UNet based architecture to autoregressively generate future frames using the refined optical flow map by forward-warping the input image features at different resolutions. We conduct extensive experiments on a publicly available dataset and show that our method is superior to the baselines in terms of qualitative and quantitative metrics. In addition, we show the qualitative animations of the objects in directions that did not exist in the training set and provide a way to synthesize videos that otherwise would not exist in the real world. Project url: \url{https://controllable-cinemagraphs.github.io/}
\end{abstract}

\section{Introduction}
It is widely perceived that animations capture human imagination more than still images. The effect of this can be seen in the proliferation of video content that is being uploaded on social media. Studies show that video-based ads and explainers are far more likely to gain trust and engagement than those based on other modalities, leading to a significant boost in sales. However, the required animations or videos are less easily available for the users to leverage than still images that exist in abundance in one's collection. Hence, it is desirable to empower the practitioners with controllable tools to convert the still images to videos of the required kind. This motivates us to consider the problem of animating images with user control to generate output videos that are generally called `cinemapgraphs' in literature. Similar to \cite{Holynski_2021_CVPR} we focus on the images that contain fluid elements like water, smoke, fire that have repeating textures and continuous fluid motion. 
\\
There has been a rich body of work \cite{halperin2021endless, blattmann2021ipoke, hao2018controllable, reda2018sdc, minderer2019unsupervised, castrejon2019improved, Holynski_2021_CVPR, dorkenwald2021stochastic, villegas2019high, franceschi2020stochastic} that has focused on generating animations from still images. While \cite{villegas2019high, reda2018sdc, castrejon2019improved, Holynski_2021_CVPR, minderer2019unsupervised, franceschi2020stochastic} focus on uncontrollable image-to-video synthesis, attempts \cite{halperin2021endless, blattmann2021understanding, blattmann2021ipoke, hao2018controllable, dorkenwald2021stochastic} have been made for controllable image-to-video synthesis with the user-provided direction of the motion of the objects in the images. While these methods provide some control to the user, they suffer from certain drawbacks. Specifically, \cite{dorkenwald2021stochastic, blattmann2021understanding, blattmann2021ipoke} either allow the user to poke at just a single pixel location or provide a single user direction. Halperin et. al \cite{halperin2021endless} obtain a displacement field by exploiting self-similarity that exists in images of repeating structures like buildings, staircases. Howeve such a method is unsuitable for animating fluid objects that we are considering as such objects do not have specific structures that can manifest in self-similarity, leading to an erroneous displacement field. Hao et. al \cite{hao2018controllable} proposed a method in which the user can provide sparse trajectories as input, defined by the direction of the motion at different locations. A dense optical flow map is estimated in an unsupervised manner and is warped with the input image to obtain the future frames. As shown later, the dense optical flow estimated thus is brittle and is prone to produce unrealistic video synthesis results. Motivated by this, we consider the problem of animating images given i) a single still image ii) a user-provided mask specifying the region to be animated and iii) a set of movement directions, called flow hints at different locations in the masked region. 
\\
To circumvent the problems associated with directly obtaining a flow map from user inputs, we propose a two-step approach to estimate the flow map from a sparse set of arrow directions and their associated speeds. Firstly, we approximate the dense optical flow using simple exponential operations on the movement directions and speeds input by the user. Next, the thus estimated approximate flow map is further refined using a GAN-based network \cite{goodfellow2014generative} to obtain the final estimate of the flow map representing the constant 2D flow map of the desired movement. The estimated flow map along with the input image is fed into a GAN-based image generator, similar to Holynski et al. \cite{Holynski_2021_CVPR} to obtain the future frames.
The contributions of our paper are as follows.
\begin{itemize}
    \item We propose a two-stage approach to interactively control the animation of fluid elements from a still image.
    \item We propose a novel approach to approximate the constant flow map governing the motion using simple exponential operations on the user-provided inputs in the form of speed and directions.
    \item Through qualitative and quantitative experiments, we show that our method beats all previous and other proposed baselines on a publicly available dataset of images of fluid motion.
    \item We prove the generalizability of our method to any arbitrary set of user directions by showing the qualitative animations of fluid objects in directions that did not exist in the training set.
\end{itemize}

\section{Related Work}
Video synthesis works occur in a myriad of ways. A good number of works have focused on video synthesis in an unsupervised and stochastic manner \cite{villegas2019high, reda2018sdc, minderer2019unsupervised, franceschi2020stochastic, castrejon2019improved, villegas2017decomposing, wu2020future}. There is a body of research that deals with video generation from intermediate representations like semantic label map \cite{pan2019video, wang2018video, wang2019few}. Of relevance to this work are the works on single-image-to-video synthesis \cite{chuang2005animating, xiong2018learning, li2018flow, wu2020future, logacheva2020deeplandscape, zhang2020dtvnet, halperin2021endless, dorkenwald2021stochastic, blattmann2021understanding, blattmann2021ipoke, hao2018controllable, endo2019animatinglandscape, Holynski_2021_CVPR}. Chuang et al. \cite{chuang2005animating} animate pictures by allowing the users to decompose the images into several layers, each one of which being needed to be animated in a different fashion. Xiong et al. \cite{xiong2018learning} propose a two-stage approach to synthesize a video from a single image, wherein in the first stage a sequence of frames is generated using a 3D-GAN and in the second stage, the sequence of frames are further refined using another GAN. Li et al. \cite{li2018flow} first predict a sequence of optical flow maps for future frames from the input image and then use them to obtain the future RGB frames. Logacheva et al. \cite{logacheva2020deeplandscape} propose a radically different approach by modeling the sequence of landscape frames in a video in the StyleGAN \cite{karras2020analyzing} latent space while enforcing the temporal consistency. Similar to \cite{li2018flow}, Holynski et al. \cite{Holynski_2021_CVPR} first estimates the optical flow for future frames, except that the work assumes a constant 2D flow map across the video. These methods \cite{xiong2018learning, li2018flow, logacheva2020deeplandscape, li2018flow, Holynski_2021_CVPR, chuang2005animating} generate a video from a single still image automatically and thus do not allow the user interaction to control the animation.
\\
Different from the above set of works, \cite{halperin2021endless, dorkenwald2021stochastic, blattmann2021understanding, blattmann2021ipoke, hao2018controllable,  endo2019animatinglandscape} allow the users to interact and control the movement in the animation to varying degrees, and hence are more closely related to the current work. Dorkenwald et al. \cite{dorkenwald2021stochastic} propose a one-to-one mapping between image and video using a residual representation, that allows the user to provide a single direction of motion for video generation. \cite{blattmann2021ipoke} and \cite{blattmann2021understanding} propose methods that govern the animation of different parts in the image with a single poke at a particular location defined by the start and end location of the motion. However, these methods \cite{dorkenwald2021stochastic, blattmann2021ipoke, blattmann2021understanding} are unsuitable for our problem that necessitates the use of a sparse set of input directions and speeds at arbitrary locations. The closest approach to our work is \cite{hao2018controllable}. This approach allows for user interaction through sparse trajectories for every frame to be predicted. Given the sparse trajectory for a particular frame, a single network is used to obtain a dense optical flow map and a hallucinated image. The dense optical map is bilinearly warped with the input image to obtain an estimate for the frame that is further added to the hallucinated image to obtain the final predicted frame. We differ from this method in two ways, i) instead of obtaining the dense optical flow directly from sparse trajectories, we first obtain its approximation by applying simple exponential functions on the user inputs, and then refining it using a network. and ii) instead of obtaining the final image by simple bilinear interpolation on the input image, we adapt the method from \cite{Holynski_2021_CVPR} and use a separate network that takes in the input image and flow map corresponding to a particular frame, with symmetric splatting of intermediate features to obtain the RGB frame.

\section{Methodology}

\begin{figure*}[t]
\begin{center}
\includegraphics[width=\linewidth]{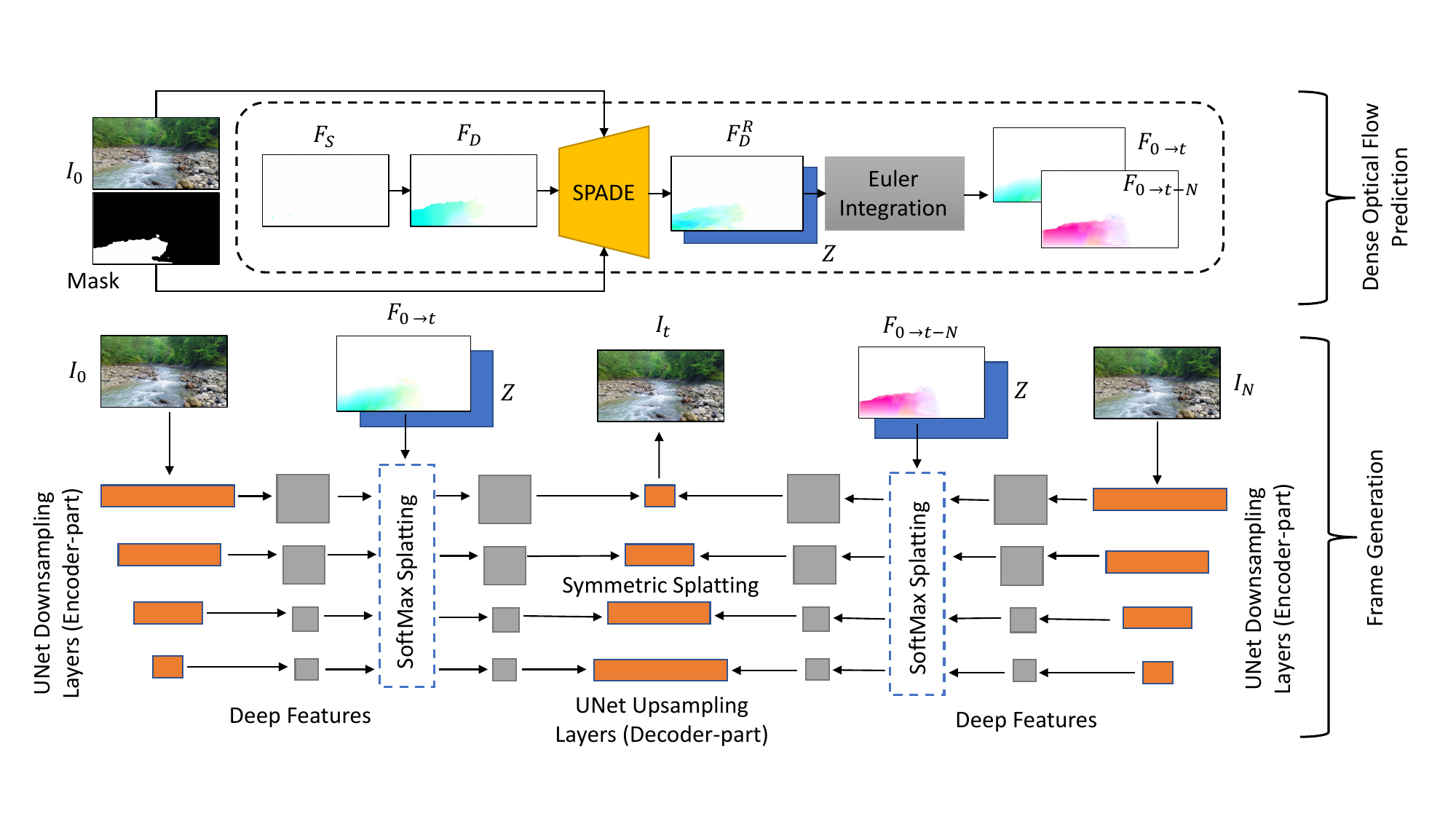}
\end{center}
\vspace{-1 cm}
   \caption{The figure shows our full pipeline. The inputs to our system are the input image, the user-provided mask indicating the region to be animated, and motion hints, $F_S$. The motion hint is converted into a dense flow map $F_D$ using simple exponential operations on $F_S$, which is further refined using a SPADE network, $G_F$ to obtain $F^{R}_D$. During test time, instead of using both $I_0$ and $I_N$, we obtain the $t^{th}$ frame, $I_t$ as the output of the UNet into which we feed the input image in place of both $I_0$ and $I_N$ and the Euler integrated flow maps corresponding to that frame, in both forward and backward directions that are used to perform symmetric splatting in deep feature space.}
\label{fig:model}
\end{figure*}

Given a single input image $I \in \mathbb{R}^{H \times W \times 3}$, mask of the regions in the image the user wants to animate $M \in \mathbb{R}^{H \times W \times 1}$, $K$ arrow directions $A^{1..,K}$ and corresponding speed values $S^{1..,K}$, our goal is to generate a realistic animated video comprising of $N$ frames ($I^{1..,N}$). Our method consists of first converting the arrow directions and speed values to sparse input flow maps, $F_S \in \mathbb{R}^{H \times W \times 2}$ (Section \ref{sparse}), where flow at position $i$ (at the location of arrows in $F_S$) defines the position where a pixel at that location will move to in all future frames. We propose to convert $F_S$ into a dense flow map $F_D$ (Section \ref{dense}) using simple exponential functions. Using $F_D$ and input image $I$ as guidance, we use a flow-refinement network to generate a dense refined flow map $F^{R}_D$ (Section \ref{refined}). Finally, to generate output video frames, we use a UNet \cite{ronneberger2015u} based frame generator to generate video frames by warping the input image with $F^{R}_D$ at different resolutions of feature maps (Section \ref{video}).

\subsection{Preliminary: Eulerian Flow Fields} \label{preliminary}
Generating video frames by warping a single image using optical flow requires very accurate optical flow to obtain realistic video. In the case of real-world videos, the optical flow between each pair of frames in a video, in most cases, is time-varying. Following this principle, Endo et al. \cite{endo2019animatinglandscape} predicts optical flow autoregressively using the previously generated frame. Although theoretically, this seems feasible, in the long-term it leads to large distortion due to error propagation. In contrast, Holynski et al. \cite{Holynski_2021_CVPR} hypothesize that a constant and time-invariant optical flow field $M_F$, termed as Eulerian flow, that describes the motion of pixel locations between consecutive frames in a video, can accurately approximate the complex motion of fluid elements (like water, smoke, fire, etc.) in realistic videos. More specifically, for a given pixel location $i$, the optical flow $F_{t \to t+1}$ between consecutive frames in a video at any time $t$ is given by,
\begin{equation}
\label{eq:eularian_1}
\begin{split}
    F_{t \to t+1}(i) = M_F(i) 
\end{split}
\end{equation}
Correspondingly, the optical flow between the first frame and frame at any time $t$, can be obtained by Euler-integration of $M_F$, $t$ times as given by,
\begin{equation}
\label{eq:eularian_2}
\begin{split}
    F_{0 \to t}(i) = F_{0 \to t-1}(i) + M_F(i+F_{0 \to t-1}(i)) 
\end{split}
\end{equation}
where $F_{0 \to 1} = M_F$.

Since we also operate in the domain of fluid elements, we adopt the principle proposed in \cite{Holynski_2021_CVPR} to use a constant optical flow field to model the motion of elements in the generated frames.

\begin{figure}[t]
\begin{center}
\includegraphics[width=\linewidth]{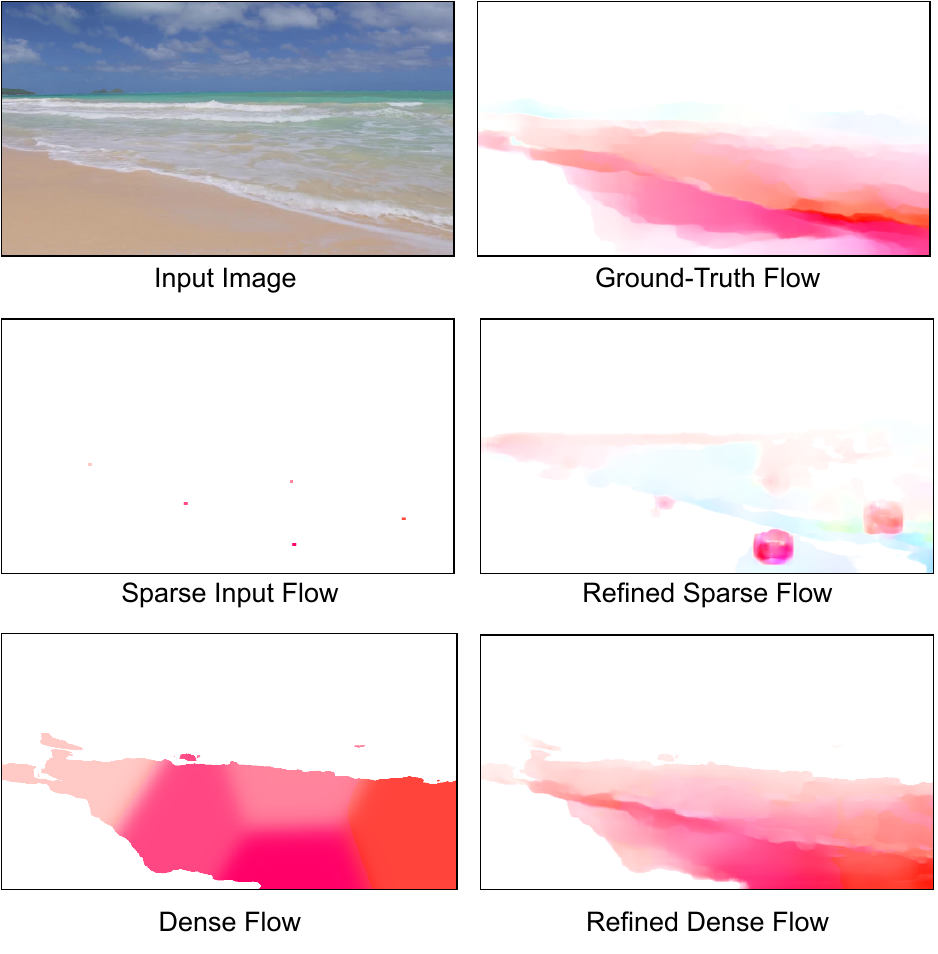}
\end{center}
   \caption{The figure shows the input image, the sparse input flow in the form of flow hints and the various flows that can be obtained. It is clear that that dense flow computed using simple functions closely approximates the ground-truth flow, and the refined dense flow best resembles the ground-truth flow among all. The refined sparse flow obtained is quite poor.}
\label{fig:flows}
\end{figure}

\subsection{Baseline: Sparse Input Flow $F_S$ $\to$ Sparse Refined Flow $F^{R}_S$} \label{sparse}
We convert the arrow directions $A^{1..,K}$ and corresponding speed $S^{1..,K}$ into sparse optical flow map $F_S$. Each arrow $A^n$ (where $n \in [0,K]$) at location $i$ in the image, is given by the start $(i)$ and end $(j)$ location of the arrow $(x^i,y^i) \to (x^j,y^j)$. The sparse optical flow $F_S$ is formulated as,
\begin{equation}
\label{eq:sparse}
  F_S(x^i,y^i) =
    \begin{cases}
      (x^j,y^j)*S^n & \text{if $A^n$ starts at $(x^i,y^i)$}\\
            0 & \text{otherwise}
    \end{cases}       
\end{equation}
Following Hao et al. \cite{hao2018controllable}, who generate time varying dense flow maps from input sparse trajectories, we transform the sparse flow map $F_S$ to a dense optical flow map $F^{R}_S$. However, unlike their method that rely on generation of time-varying flow maps, we generate a constant (Eulerian) dense optical flow $F^{R}_S$. For this we use a SPADE \cite{park2019semantic} based flow-refinement network $G_F$ by using $F_S$, input image and mask as cues, in the SPADE normalization layers. We call this baseline to compute dense flow map as `Hao et al. modified'.

\subsection{Sparse Input Flow $F_S$ $\to$ Dense Flow $F_D$} \label{dense}
From fig. \ref{fig:flows}, it can be observed that the refined dense optical flows $F^{R}_S$ generated using $F_S$ are very different from ground-truth Eulerian flows. We hypothesize that $F_S$ does not provide adequate information to $G_F$ to produce realistic flows. Instead of directly using $F_S$ to generate dense flow using $G_F$, we propose to create an intermediate dense optical $F_D$ from $F_S$ using simple functions of the distances between every pixel location and the arrow positions. We calculate the exponential $L2$ Euclidean distance $(D_{exp})$ between each pixel location in input image and the starting coordinate of all the K arrows. The exponential of the Euclidean distance between location $(x^i,y^i)$ of input image and starting position of arrow $A^j$, $(x^j,y^j)$ is given by,
\begin{equation}
\label{eq:gs_flow_1}
  D_{L2}^{i,j} = \|(x^i,y^i)\ -(x^j,y^j)\|_2     
\end{equation}
\begin{equation}
\label{eq:gs_flow_2}
  D_{exp}^{i,j} = e^{-\left(D_{L2}^{i,j}/\sigma\right)^2} 
\end{equation}
where $D_{L2}^{i,j}$ is the Euclidean distance between location $(x^i,y^i)$ of input image and starting position of arrow $A^j$, $(x^j,y^j)$ and $\sigma$ is a constant. The dense optical flow $F_D$ for a particular pixel location $i$ in in the image is defined as weighted average of flow magnitude at each non-zero location in input mask $M$, where the weights are taken from $D_{exp}^{i,j}$ and is given by,
\begin{equation}
\label{eq:gs_flow_3}
  F_D(i) =
    \begin{cases}
        \frac{\sum_{j=1}^{K}D_{exp}^{i,j}*F_{S}(j)}{\sum_{j=1}^{K}D_{exp}^{i,j}} & \text{if $i \in M$}\\
        0 & \text{if $i \notin M$}
    \end{cases}
\end{equation}

\subsection{Dense Flow $F_D$ $\to$ Dense Refined Flow $F^{R}_D$} \label{refined}
Although $F_D$ can suitably describe the motion of fluid regions moving approximately in the same direction, and also having smooth transition at the boundary of two different flow regions, there are limitations of directly using $F_D$ to generate video frames by warping. In the figures \ref{fig:flows} and \ref{fig:better}, we consider examples and show the approximated dense flow and the refined dense flow. While the dense flow is a good approximation, the refined dense flow is closer to the ground-truth flow than the approximated flow. From figure \ref{fig:better} we see that the dense flow map generated is assigned the same horizontal flow to the majority of the waterfall and below the lake, whereas, realistically the waterfall would have been moving vertically downward. This is due to the fact the $F_D$ is generated purely based on the closeness of a particular pixel location to hint points with simple exponential operations. Hence, unlike $F^{R}_D$, it cannot distinguish the object boundaries (in this case the dotted demarcation like between waterfall and lake in figure \ref{fig:better}) and naively mixes the flow values in both regions. Due to this same drawback, it assigns very different flow values to different regions of the same waterfall, which again is realistically inappropriate. Hence, using our flow-refinement network $G_F$, with the input image, mask, and $F_D$ as cues, we generate a dense refined optical flow $F^{R}_D$. Using $F^{R}_D$ as Eulerian flow fields, we generate video frames by the method discussed in Section \ref{video}.

\subsection{Video Frame Generation} \label{video}
From the refined dense optical flow field $F^{R}_D$, we estimate the flow fields $F_{0 \to t}$ from input image to all the future frames for $t\in [0,N]$ using equation \ref{eq:eularian_2}. Instead of the backward optical flow field (as used in \cite{endo2019animatinglandscape}), we use forward flow to warp input frame to generate future frames as \cite{Holynski_2021_CVPR} observed that forward flow produces more reliable flow estimates and sharper values at object boundaries. However, forward warping (also known as splatting) has its challenges \textit{(i)} it can map multiple source pixels to the same destination resulting in loss of information and aliasing, \textit{(ii)} it may not map any source pixel to a particular target location leading to blank region. To mitigate these artifacts, we use softmax-splatting (proposed in \cite{Niklaus_CVPR_2020}). It resolves the first challenge associated with splatting by using softmax to scale the contribution of source pixels mapping to the same destination location, based on importance metric $Z \in \mathbb{R}^{H \times W \times 1}$. In our method, we predict $Z$ as an additional channel in the output of $G_F$ during fine-tuning (Section \ref{fine-tuning}).

\subsubsection{Multi-scale Deep Feature Warping} \label{warping}
Instead of directly splatting the input image with $F_{0 \to t}$ in RGB space to generate frames, that might otherwise produce holes in generated frames, we perform splatting on the deep features of the image, similar to \cite{Holynski_2021_CVPR}. However, unlike their method, we perform splatting at different resolutions of input image features. Using a UNet based image generator $G_I$, we extract the image features at different resolutions from the UNet encoder. We use softmax-splatting to warp the features at different scales and generate an image using the decoder of the UNet. Note that all the splatted features from the encoder part, except at the bottleneck layer of UNet, are connected to the decoder via skip connections. For a feature map corresponding to input image $I$ at resolution $D_{0}^r$, the softmax-splatting output $D_{t}^r$ at pixel location $i$ using $F_{0 \to t}$ is given by,
\begin{equation}
\label{eq:softmax-splatting-1}
  D_{t}^r(i) = \frac{\sum_{j \in \mathcal{X}}D_{0}^r(j)e^{Z(j)}}{\sum_{j \in \mathcal{X}}e^{Z(j)}}
\end{equation}
where $\mathcal{X}$ consists of all the pixels that map to the same target location $i$ after splatting.

\subsubsection{Symmetric Splatting} \label{symmectric-splatting}
Even with softmax-splatting on  multi-scale image feature space, we observe the presence increasingly large void regions, in the generated frames (similar to what was observed in \cite{Holynski_2021_CVPR}) in places of significant motion where pixels are warped away from the region and there are no pixels to replenish them. We hypothesize that $G_I$ might not be able to generate appropriate pixel values for these regions to fill in the gaps. To resolve this artifact we use the method of symmetric-splatting proposed in \cite{Holynski_2021_CVPR}. In this method, similar to producing flow fields in the forward direction $F_{0 \to t}$, we also generate flow fields in the backward direction $F_{0 \to t-N}$ by Euler-integration of $-M_F$. Thus, instead of just using softmax-splatting on deep features $D_{0}^r$ obtained from first frame $I_0$ with $F_{0 \to t}$ to generate $D_{t}^r$, we use a combination of the deep features $D_{0}^r$ from $I_0$ and $D_{N}^r$ obtained from last frame $I_N$, splatted with $F_{0 \to t}$ and $F_{0 \to t-N}$ respectively. Specifically, any given pixel location $i$ in the combined deep feature $\hat{D}_{t}^r$ is given by,
\begin{equation}
\label{eq:softmax-splatting}
  \hat{D}_{t}^r(i) = \frac{\sum_{j \in \mathcal{X}}\alpha D_{t}^r(j)e^{Z(j)}+\sum_{\hat{j} \in \mathcal{\hat{X}}}\hat{\alpha}D_{t-N}^r(\hat{j})e^{Z(\hat{j})}}{\sum_{j \in \mathcal{X}}\alpha e^{Z(j)}+\sum_{\hat{j} \in \mathcal{\hat{X}}}\hat{\alpha}e^{Z(\hat{j})}}
\end{equation}
where $\alpha$ and $\hat{\alpha}$ equals $(1-\frac{t}{N})$ and $\frac{t}{N}$, $D_{t}^r(i)$ and $D_{t-N}^r(i)$ are feature map obtained by softmax-splatting $D_{0}^r$, $D_{N}^r$ with $F_{0 \to t}$ and $F_{0 \to t-N}$ respectively, $\mathcal{X}$ and $\mathcal{\hat{X}}$ consists of all the pixels that map to the same target location $i$ after splatting for $D_{t}^r$ and $D_{t-N}^r$ respectively. 
The intuition behind this is the void regions appearing in the frames generated at time $t$ by splatting the first image with $F_{0 \to t}$ is complementary to the void regions appearing in the frames generated by splatting the last image with $F_{0 \to t-N}$.

\subsubsection{Training and Inference} \label{fine-tuning}
As proposed by \cite{Niklaus_CVPR_2020}, for stable training, we first train the 2 components of dense optical flow refinement and frame generator separately. While training $G_F$, we use the standard GAN loss and Discriminator Feature-matching loss \cite{park2019semantic}. In this stage, we compute losses based on generated dense refined optical flow (and not the generated $Z$). During the training of $G_I$, we use the standard GAN loss, VGG loss \cite{simonyan2014very}, L1 loss, and Discriminator Feature-matching loss. Prior to end-to-end fine-tuning, we freeze the refined dense optical flow maps and only train $G_F$ to generate $Z$. In addition, we only use the discriminator for the frame generator part. Contrary to training where we use both the first and the last frames for symmetric-splatting used in $G_I$, at test time, since we only have a single input static image, use the same image as both the first and the last frames in $G_I$.

\begin{figure*}[t]
\begin{center}
\includegraphics[width=\linewidth]{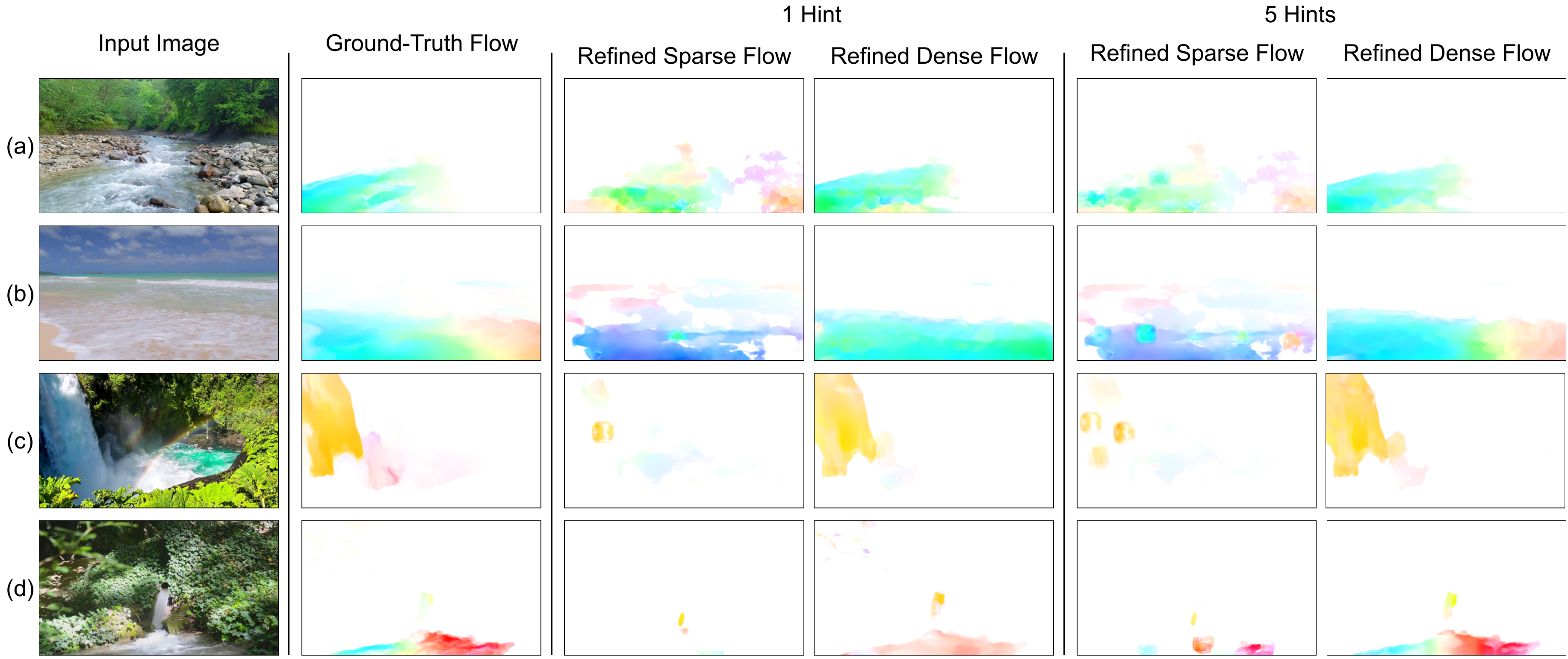}
\end{center}
\vspace{-0.4cm}
   \caption{The figure shows the comparison of the refined sparse flow and the refined dense flow for one hint and five hints. It is quite evident that across diverse set of hints, the refined dense flow resembles the ground-truth flow far more closely that refined sparse flow.}
\label{fig:flow_hints}
\end{figure*}

\begin{figure}[t]
\begin{center}
\includegraphics[width=\linewidth]{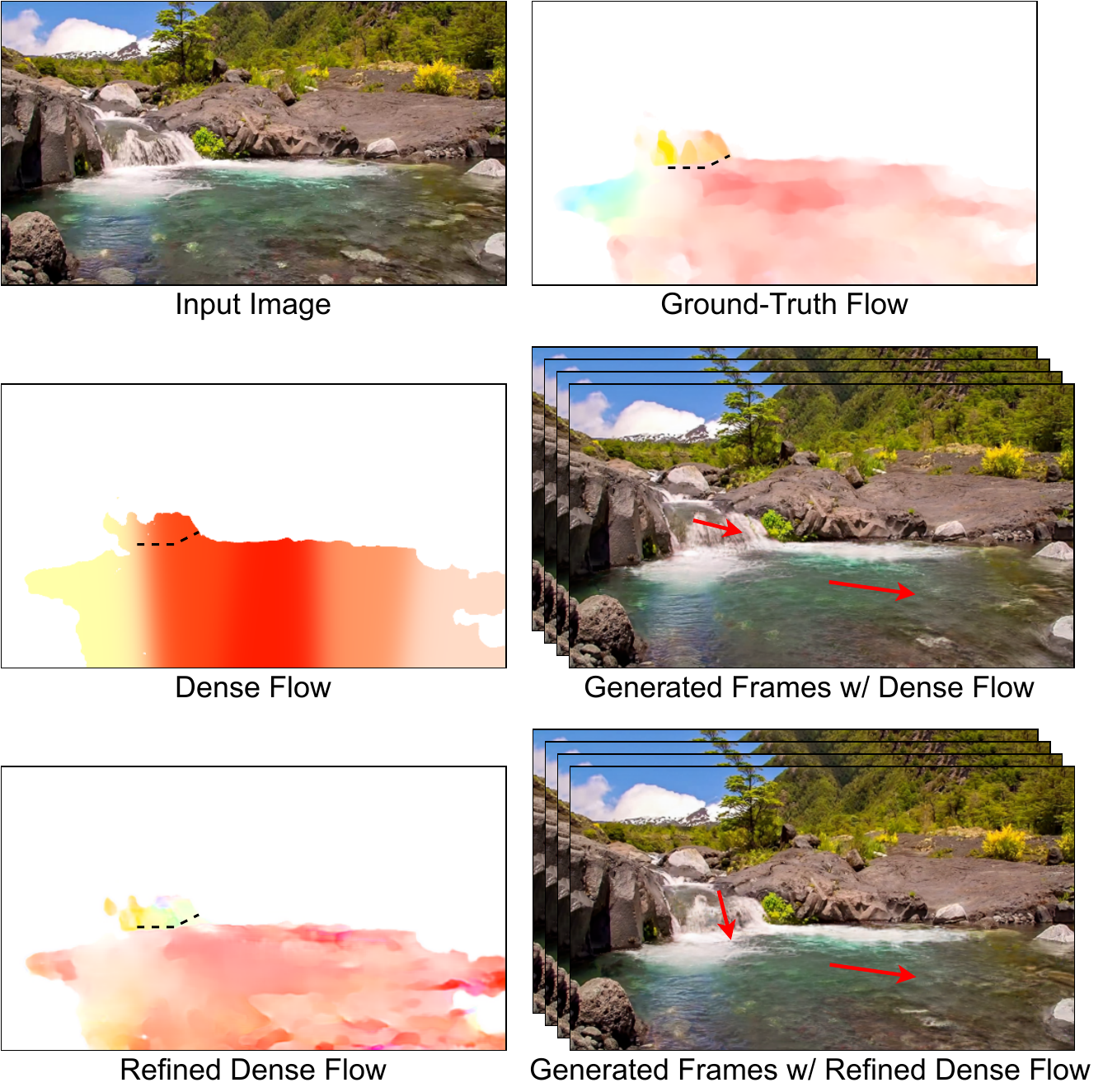}
\end{center}
   \caption{The figure shows the dense flow and the refined dense flow for a particular example and the corresponding generated frames of the video. While the dense flow is a good approximation, the direction of motion is more appropriate in the video corresponding to the refined dense flow that closely resembles the ground-truth flow. In the dense flow video, the waterfalls tend to shift sideways to the right, while the refined dense flow video is very similar to the ground-truth video. The videos are shown in the supplemental. The dotted line (\textbf{- - -}) denotes the region where the waterfall ends and the lake starts.}
\label{fig:better}
\end{figure}

\section{Experiments}
\subsection{Dataset preparation}
Due to the unavailability of any existing human-annotated controllable video generation datasets with masks, input flow hints/arrows in the domain of natural scenes, we curate our dataset from the uncontrollable video generation dataset provided by \cite{Holynski_2021_CVPR}. This dataset already contains the ground-truth videos, starting frame, and the average optical flow for all videos. The number of frames across all videos is 60, each frame having a resolution of 720x1280. For our purpose of training and testing, in addition to the input image, and average flow, we also require a mask of regions the user wants to annotate along with arrows and corresponding speed values. We generate a substitute of the user annotated mask and arrows with the mask and flow hint points generated heuristically from ground-truth average optical flow.

\textbf{Mask Generation:} For every average optical flow map $F_{avg}$ in the dataset, we calculate the mean-squared flow value for $F_{avg}$. Then we mask out all the regions having per-pixel squared-flow less than $m$ times the mean-squared flow value. Following observation on the visual quality of generated mask at different $m$ values, we set the value of $m$ to 10.

\textbf{Flow Hint Generation:} Using the mask $M$ generated in the previous step, we calculate masked average flow $F_{avg}^M$ for each video as $M*F_{avg}$, where $F_{avg}$ is the ground-truth average flow map for that video. We perform k-means clustering for 100 iterations on the individual $F_{avg}^M$, to find the cluster centers based on the number of desired hint points for our input sparse optical flow map $F_S$. $F_S$ consists of values equal to the ground-truth average flow maps at the pixel location of cluster centers and zero elsewhere. In our experiments, the number of hints points are chosen to be either 1, 3, or 5. In Section \ref{sparse}, we discuss the procedure to convert user-provided arrows and speed values to $F_S$, which is required in real-world interactive testing.

\subsection{Experimental Setup}
For flow refinement network $G_F$, we use SPADE \cite{park2019semantic}. We also use multi-scale discriminator $D_F$ from \cite{park2019semantic} during training. We train the flow refinement part of our method separately for 200 epochs with both generator and discriminator learning rates of $2 \times 10^{-3}$ with TTUR method of updating learning rate proposed in \cite{park2019semantic}. We train on a triplet of (first frame, $F_S$, ground-truth average flow), where $F_S$ is randomly selected to have 1, 3, or 5 flow hints.
For frame generation network $G_I$, we use a modification of UNet (shown in fig. \ref{fig:model}) which incorporates symmetric-splatting. We use the same multi-scale discriminator $D_I$ from \cite{park2019semantic} during training. We train the frame generation part separately for 200 epochs on training tuples of (start frame, ground-truth average flow, middle frame, last frame), where the middle frame is selected randomly from time $[1,59]$. Both generator and discriminator learning rates are set to $2 \times 10^{-3}$ with the TTUR method of updating the learning rate. 
During fine-tuning we only use $G_F$, $G_I$ and $D_I$. Additionally, we fix the value to $G_F$ that is responsible for flow generation and only keep $Z$ trainable. Both generator and discriminator learning rates are reduced to $1 \times 10^{-3}$. We train for 40 epochs. Prior to training, we resize all the average flows and frames to 288x512 (maintaining the $\frac{16}{9}$ aspect ratio of original frames). At inference, we generate 60 frames.

\subsection{Baselines}
We compare our final method with five different baselines, one of our own, Endo et al. \cite{endo2019animatinglandscape}, Hao et al. \cite{hao2018controllable}, and the vanilla Eulerian method \cite{Holynski_2021_CVPR}. Our own baseline is computing the dense flow map using simple exponential functions followed by an image generator. Endo et al. provide an optimization procedure to compute the directions and speed during test time. For Hao et al., we repurpose the method they have provided in their paper by making the following modifications. Instead of obtaining sparse trajectories, we use the sparse hints that were obtained using the procedure outlined earlier in the section. Instead of having a single network to convert the sparse user inputs, we first compute dense flow map directly from the sparse hints by training the same procedure, i.e the GAN-based network. Once the dense flow map is obtained, instead of bilinearly warping the input image (as done in Hao et al.), we use the same image generator that is used for our method to generate the frames from the computed dense optical flow and the input image. We dub this baseline as `Hao et al. modified + frame generator'. In addition, we also compare our results with the vanilla Eulerian method which is fully automatic and does not require any user inputs.

\subsection{Metrics}
In order to evaluate our method against the various baselines, we use the following metrics.\\
\textbf{Frechet Video Distance (FVD) \cite{unterthiner2018towards}:} It is a standard metric used to quantify the fidelity of the generated videos and provides a measure of the distance between the generated videos and the real videos. Prior to obtaining the features, we resize all videos to $224 \times 224$ and use $60$ frames. To obtain the features from the videos, we use the pre-trained I3D \cite{szegedy2016rethinking} model that was trained on Kinematics dataset \cite{kay2017kinetics}.\\
\textbf{PSNR:} While FVD assesses the perceptual quality of the generated videos, we assess the mean pixel accuracy using PSNR. Given that it is based on mean square error, PSNR tends to favor those methods that produce somewhat blurry results.

\begin{table}[t]
\begin{adjustbox}{width=0.45\textwidth}
\centering
\begin{tabular}{|c|c|c|c|}
\hline 
 & \textbf{Method} & \textbf{FVD} $\downarrow$ & \textbf{PSNR} $\uparrow$\\
\hline \hline
\multirow{4}{*}{1 Hint} 
& Endo et al. & 561.33 & 23.59\\
\hhline{~|---}
& Hao et al. modified + frame generator & 419.015 & 25.12\\
\hhline{~|---}
& Our ($F_{D}$ + frame generator) & 419.49 & \textbf{25.2}\\
\hhline{~|---}
& Our ($F_{D}^R$ + frame generator) & \textbf{380.475} & 25.07\\
\hline \hline
\multirow{4}{*}{3 Hints} 
& Endo et al. & 526.55 & 23.35\\
\hhline{~|---}
& Hao et al. modified + frame generator & 375.98 & 25.11\\
\hhline{~|---}
& Our ($F_{D}$ + frame generator) & 331.8 & \textbf{25.22}\\
\hhline{~|---}
& Our ($F_{D}^R$ + frame generator) & \textbf{318.39} & 25.09\\
\hline \hline
\multirow{4}{*}{5 Hints} 
& Endo et al. & 519.18 & 23.21\\
\hhline{~|---}
& Hao et al. modified + frame generator & 344.55 & 25.11\\
\hhline{~|---}
& Our ($F_{D}$ + frame generator) & 335.4 & \textbf{25.24}\\
\hhline{~|---}
& Our ($F_{D}^R$ + frame generator) & \textbf{315.31} & 25.1\\
\hline \hline
& Vanilla Eulerian & 419.74 & 25.2\\
\hline
\end{tabular}
\end{adjustbox}
\caption{The table shows the FVD and PSNR values for various methods that use different number of hints as well as the Vanilla Eulerian method for the original speed. It clearly shows our method that uses refined flow and frame generator performs the best in terms of FVD, thus showing the high-fidelity animations generated by our method.}
\vspace{-0.3cm}
\label{tb:metrics_1}
\end{table}

\begin{figure}[t]
\begin{center}
\includegraphics[width=\linewidth]{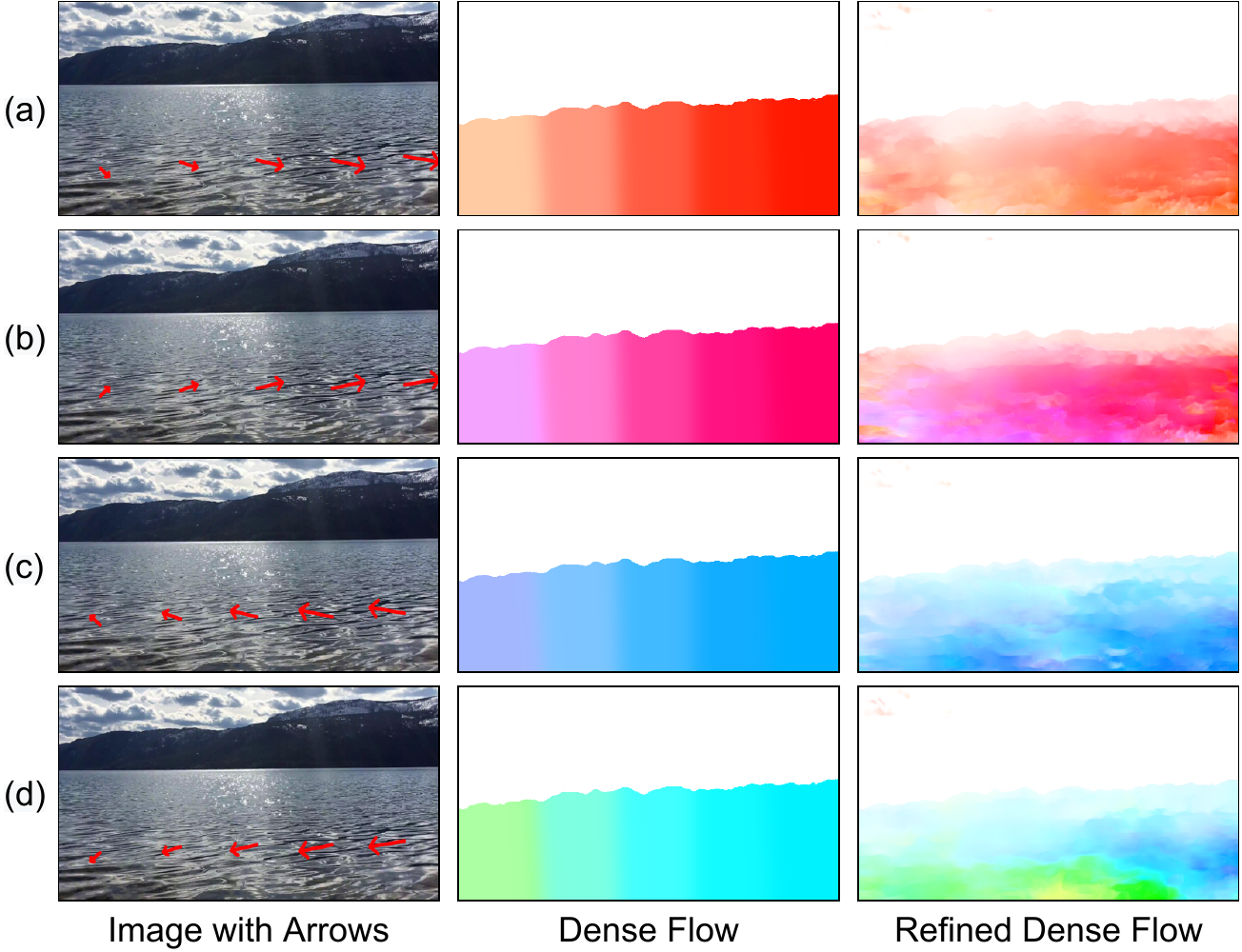}
\end{center}
   \caption{The figure shows dense flows and corresponding refined dense flow maps for four different arrow directions for the same input image. This shows the robustness of our flow generation method to arbitrary input directions for flow hints.}
   \vspace{-0.3cm}
\label{fig:controlled}
\end{figure}

\subsection{Results}
\paragraph{Evaluating the flow map generation:}
In order to show the efficacy of our flow map generation method, we compute the mean PSNR between the generated flow maps and the ground-truth flow maps. The results for three different numbers of hints are shown in table \ref{tb:flow_psnr}. It is clear that our method that involves flow refinement network on dense flow is able to generate better flow maps than just sparse hints with refinement network, in terms of PSNR. This trend is consistent across a different number of hints. It is interesting that for the vanilla Eulerian method, the PSNR is lower than both methods. This is expected given that the Eulerian method is automatic, unlike the other two methods. Similarly, we show in figure \ref{fig:flow_hints}, for various examples and the different number of hints, that the flow maps computed by our method perform significantly better than `Hao et al. modified' and more closely resemble the ground-truth flows.
\vspace{-0.1cm}
\paragraph{Evaluating the video generation:}
Table \ref{tb:metrics_1} shows the comparison of the various methods for two different metrics for three different numbers of hints. It is clear that our methods beat all the baselines in terms of FVD, while the PSNR is comparable across the different methods except Endo et al is the least of all. We also observe that our method (Our ($F_{D}$ + frame generator)) is better than `Hao et al. modified', thus showing the need to approximate the dense flow map. In addition, our final method (Our ($F_{D}^R$ + frame generator)) is better than just (Our ($F_{D}$ + frame generator), thus buttressing the need for refining the approximated dense flow. It is also observed that the FVD scores for all methods get progressively better with an increasing number of hints.

\begin{table}[!t]
\centering
\begin{adjustbox}{width=0.3\textwidth}
\begin{tabular}{|c|c|c|}
\hline 
 & \textbf{Method} & \textbf{PSNR} $\uparrow$\\
\hline \hline
\multirow{2}{*}{1 Hint} 
& Hao et al. modified & 20.48\\
\hhline{~|--}
& Our ($F_{D}^R$) & \textbf{24.15}\\
\hline \hline
\multirow{2}{*}{3 Hint} 
& Hao et al. modified & 21.06\\
\hhline{~|--}
& Our ($F_{D}^R$) & \textbf{25.53}\\
\hline \hline
\multirow{2}{*}{5 Hint} 
& Hao et al. modified & 21.4\\
\hhline{~|--}
& Our ($F_{D}^R$) & \textbf{25.82}\\
\hline \hline
& Vanilla Eulerian & 18.28\\
\hline
\end{tabular}
\end{adjustbox}
\caption{The table shows the average PSNR between the ground-truth flows and the two refined flows. The refined dense flow is significantly better than the refined sparse flow, thus corroborating the need for the exponential operations before refinement.}
\label{tb:flow_psnr}
\end{table}

\vspace{-0.1cm}
\paragraph{Qualitative Results:} \label{qualitative}
From figures \ref{fig:flows} and \ref{fig:flow_hints}, it is clearly visible that dense refined flow $F^{R}_D$ resembles the average ground-truth optical flow much more accurately than the sparse refined flow $F^{R}_S$. The generated videos using our methods and baseline are provided in supplementary. In the figure \ref{fig:better}, we consider a particular example that shows the approximated dense flow, the refined dense flow, and a frame of the generated video (video provided in supplementary). While the dense flow is a good approximation in many situations, the refined dense flow corrects some of the regions representing inappropriate flows in dense flow map (see figure \ref{fig:better}). This is directly reflected in the quality of the animation that is generated. The video generated using the dense flow has artifacts wherein the waterfalls tend to shift progressively towards the right (due to dense flow having the same flow values for waterfall and lake), whereas the video generated using the refined flow is very realistic and resembles the actual downward motion of waterfall observed in real-world videos.

\paragraph{Animation in arbitrary directions:}
Our method is capable of generating flow maps from flow hints that correspond to any arbitrary directions that may not have existed in the training set. Figure \ref{fig:controlled} shows the refined dense flows generated using the same input image with different arrow directions, and thus different dense flows. We see that $G_F$ produces results that respect the input arrow directions and are not just based on the input image, showing the robustness and the generalizability of our method. Please see the supplement material for corresponding generated videos.

\section{Conclusions and Limitations} \label{limitations}
We propose a method to animate images that contain fluid elements like water, fire, smoke, given a user-provided mask and flow hints in the forms of speed and direction. We proposed a simple yet powerful method to approximate the constant flow field governing the motion with simple exponential operations on the user-provided flow hints, and further show that in order to obtain a better flow field we need to refine using a network the approximation for dense flow field rather than just the sparse hints. Through quantitative experiments, we show that our method performs better than all baselines for a various number of motion hints. One of the limitations of our method is that it is restricted to the movements of fluid elements in an image. The motion of rigid objects or even definite structures like designs in buildings cannot be modeled using a constant flow field, thus making our method not applicable in such scenarios. Another potential limitation of our method is its inability to model multiple flow streams that are adjacent to each other but may belong to different objects.

\clearpage
{\small
\bibliographystyle{ieee_fullname}
\bibliography{egbib}
}

\clearpage
\appendix
\noindent{\Large\bf Appendix}
\vspace{5pt}

\section{DTFVD scores:}
Similar to Dorkenwald et al, \cite{dorkenwald2021stochastic}, we use the model trained on the DTDB dataset \cite{hadji2018new} to extract features to compute the Dynamic Texture Frechet Video Distance (DTFVD) scores. As shown in the table \ref{tb:dt_fvd}, our method performs the best in terms of quality of videos generated.
\begin{table}[!t]
\centering
\begin{adjustbox}{width=0.45\textwidth}
\begin{tabular}{|c|c|c|c|}
\hline 
 \textbf{Hao et al. modified + frame generator} & \textbf{Our ($F_{D}$ + frame generator)} & \textbf{Our ($F_{D}^R$ + frame generator)}\\
\hline \hline
0.582 & 0.571 & \textbf{0.512}\\
\hline
\end{tabular}
\end{adjustbox}
\caption{The table shows the DTFVD scores to compare our method with other baselines. All the results are calculated for 5 hints.}
\label{tb:dt_fvd}
\end{table}

\begin{table*}[t]
\centering
\begin{adjustbox}{width=\textwidth}
\begin{tabular}{|c|c|c|c|c|}
\hline 
 & \textbf{Question} & \textbf{Hao et al. modified + frame generator)} & \textbf{Our ($F_{D}$ + frame generator)} & \textbf{Our ($F_{D}^R$ + frame generator)}\\
\hline \hline
\multirow{4}{*}{3 Hint} 
& \textit{Which one among three generated videos is} & \multirow{2}{*}{60} & \multirow{2}{*}{62} & \multirow{2}{*}{\textbf{128}}\\
\hhline{~|~~~~}
& \textit{more representative of the ground-truth video?} &  &  & \\
\hhline{~|----}
& \textit{Which of the three generated videos is most representative of the Direction} & \multirow{2}{*}{73} & \multirow{2}{*}{72} & \multirow{2}{*}{\textbf{105}}\\
\hhline{~|~~~~}
& \textit{of Movement as shown by arrows in image with arrow directions?} &  &  & \\
\hline \hline
\multirow{4}{*}{5 Hint} 
& \textit{Which one among three generated videos is} & \multirow{2}{*}{57} & \multirow{2}{*}{58} & \multirow{2}{*}{\textbf{135}}\\
\hhline{~|~~~~}
& \textit{more representative of the ground-truth video?} &  &  & \\
\hhline{~|----}
& \textit{Which of the three generated videos is most representative of the Direction} & \multirow{2}{*}{69} & \multirow{2}{*}{56} & \multirow{2}{*}{\textbf{125}}\\
\hhline{~|~~~~}
& \textit{of Movement as shown by arrows in image with arrow directions?} &  &  & \\
\hline
\end{tabular}
\end{adjustbox}
\caption{We show the aggregate the number of preferences for the different methods across 250 individual responses per question and per number of hints. It is clear that our method is preferred more than other methods by a significant margin.}
\label{tb:amt}
\end{table*}

\section{Human Evaluation}
We conduct human evaluation of compare the quality of our results using Amazon Mechanical Turk (AMT). The AMT workers are provided with the input images with arrow directions, the mask of the region to animate the the actual ground-truth videos along with videos generated by three different methods, Hao et al. modified + frame generator, Our ($F_{D}$ + frame generator) and Our ($F_{D}^R$ + frame generator). Each AMT worker is asked to answer two questions, \textit{i)}  Which one among three generated videos is more representative of the ground-truth video? \textit{ii)} Which of the three generated videos is most representative of the Direction of Movement as shown by arrows in image with arrow directions? Out of the 162 examples in test set, we randomly select 50 examples for two different numbers of hints (3 hints and 5 hints) accounting for a total of 100 video comparisons. Each question was answered by five AMT workers (total of 500 individual responses). 
We show the aggregate the number of preferences across 250 individual responses per question and per number of hints in table \ref{tb:amt}. As seen from the table, our method is overwhelmingly favoured for both the questions and both number of hints.

\section{Video results:}
Please refer to the links, flows and videos present in the project page \url{https://controllable-cinemagraphs.github.io/} for additional flow and video results. In particular, we would like to refer the reader to the view the controllable\_variations link where for the same input image, we show output videos corresponding to animations from different arrow directions or flow hints. 

\section{Negative societal impact of the work:}
The video synthesis framework proposed in the paper will empower the users to generate videos from a single images by making the fluids move in any arbitrary directions. This could lead to generation of videos of the kind that do not exist in real world. The users can potentially misuse to create unreasonable and unrealistic marketing ads, thus impacting the society at large in a negative manner.

\end{document}